%% file: main.tex
\documentclass[sigconf]{acmart}

\usepackage{latexsym}
\usepackage{graphicx}
\usepackage{adjustbox}
\usepackage{booktabs}
\usepackage{multirow}
\usepackage{color, colortbl}
\definecolor{LightCyan}{rgb}{0.88,1,1}
\definecolor{Gray}{gray}{0.9}

\newcommand{\ACMDarkBlue}[1]{\textcolor{ACMDarkBlue}{#1}}
\usepackage{microtype}
\usepackage{CJKutf8}
\usepackage[utf8]{inputenc}
\usepackage[T1]{fontenc}
\usepackage{tabularx}
\usepackage{algorithmic}
\usepackage{subfigure}
\usepackage{svg}

\begin{document}

\title{TAG: Toward Accurate Social Media Content Tagging\\with a Concept Graph}

\input{section/others/author}

\input{section/abstract}

\setcopyright{none}
\settopmatter{printacmref=false} 
\renewcommand\footnotetextcopyrightpermission[1]{} 
\maketitle
\pagestyle{plain} 

\input{section/intro}

\input{section/related}

\input{section/data}

\input{section/model}

\input{section/simu}

\input{section/conclude}

\bibliographystyle{ACM-Reference-Format}
\bibliography{acmart.bib}


\end{document}

%% file: section/others/author.tex









\author{Jiuding Yang$^{1*}$, Weidong Guo$^{3*}$, Bang Liu$^{2*}$, Yakun Yu$^{1}$, }
\author{Chaoyue Wang$^{3}$, Jinwen Luo$^{3}$, Linglong Kong$^1$, Di Niu$^1$, Zhen Wen$^3$}
\thanks{$^*$These authors contributed equally to this work.}
\affiliation{$^1$University of Alberta, Edmonton, AB, Canada}
\affiliation{$^2$RALI \& Mila, Universit{\'e} de Montr{\'e}al, Montr{\'e}al, QC, Canada}
\affiliation{$^3$Platform and Content Group, Tencent, Beijing, China}

\renewcommand{\shortauthors}{Yang, Guo and Liu, et al.}

%% file: section/abstract.tex
\begin{abstract}
Although conceptualization has been widely studied in semantics and knowledge representation, it is still challenging to find the most accurate concept phrases to characterize the main idea of a text snippet on the fast-growing social media. This is partly attributed to the fact that most knowledge bases contain general terms of the world, such as trees and cars, which do not have the defining power or are not interesting enough to social media app users. Another reason is that the intricacy of natural language allows the use of tense, negation and grammar to change the logic or emphasis of language, thus conveying completely different meanings. 
In this paper, we present TAG, a high-quality concept matching dataset consisting of 10,000 labeled pairs of fine-grained concepts and web-styled natural language sentences, mined from the open-domain social media. The concepts we consider represent the trending interests of online users.
Associated with TAG is a \textit{concept graph} of these fine-grained concepts and entities to provide the structural context information.
We evaluate a wide range of popular neural text matching models as well as pre-trained language models on TAG, and point out their insufficiency to tag social media content with the most appropriate concept.
We further propose a novel graph-graph matching method that demonstrates superior abstraction and generalization performance by better utilizing both the structural context in the concept graph and logic interactions between semantic units in the sentence via syntactic dependency parsing.
\end{abstract}

%% file: section/intro.tex
\section{Introduction}
\label{sec:intro}

\begin{figure}[tp]
	\centering
		    \includegraphics[width=0.87\columnwidth]{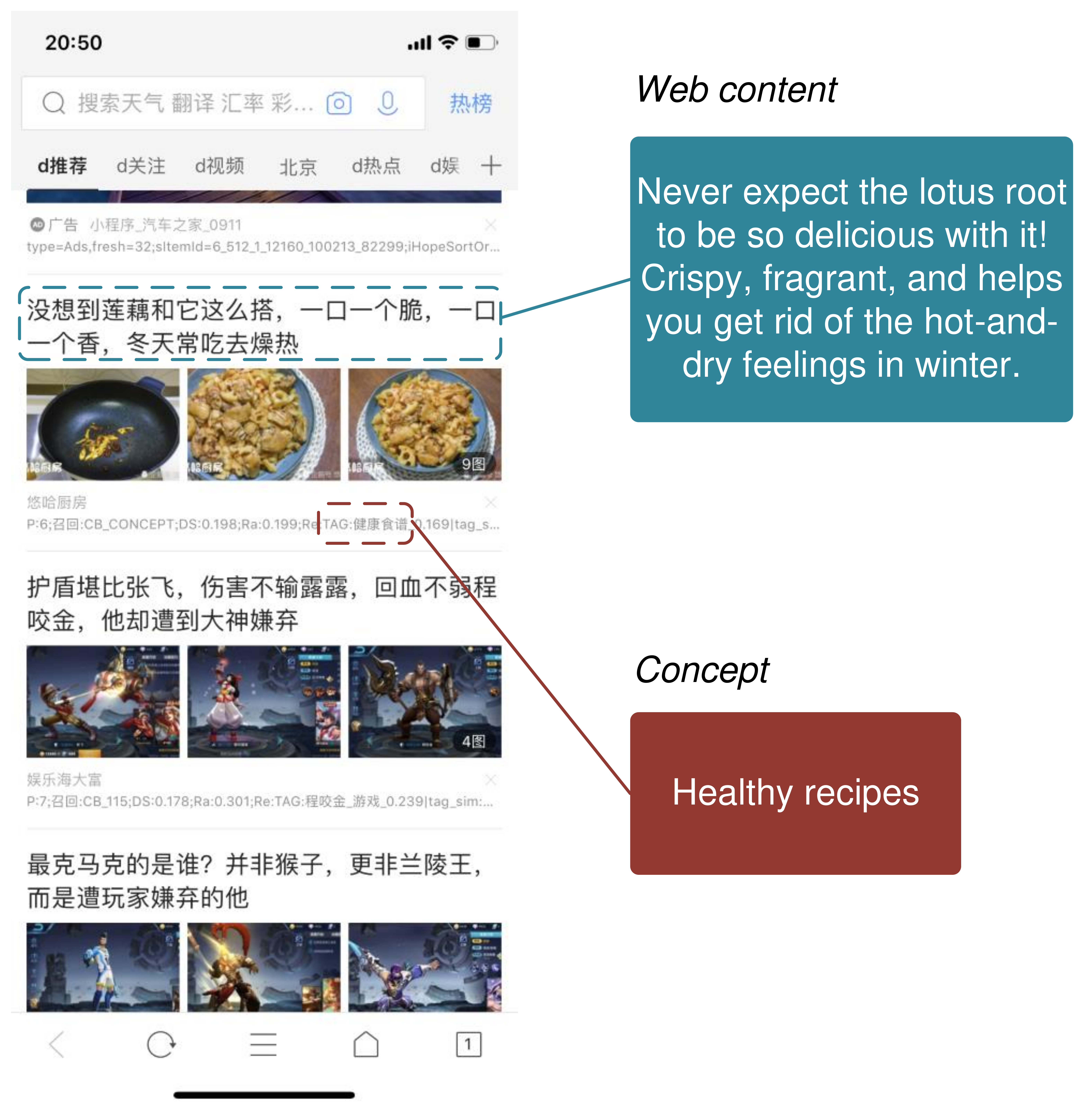}
		\caption{An example of document retrieved by concept in the news feeds stream of \href{https://browser.qq.com/}{\ACMDarkBlue{QQ Browser.}}}
	\label{case}
	\vspace{-3mm}
\end{figure}

Concepts are abstract ideas and notions about entities encountered in daily lives. 
While a common hypothesis in cognitive sciences, including linguistics, psychology and philosophy, is that all cognition must occur through concepts,
their importance is also recognized by recent research in knowledge representation and natural language processing.
Several recent efforts have focused on building a hierarchy, ontology or graph of concepts \cite{wu2012probase, liu2020giant,luo2020alicoco}, which are key to document understanding \cite{liu2019user,liu2020giant}, search query understanding \cite{song2011short,kim2013context}, as well as recommender systems \cite{liu2020giant,luo2020alicoco}.

Although many existing knowledge bases contain vast amounts of general notions for the world, e.g., trees, companies, cars, finding interesting concept terms that users would pay attention to among the exploding and fast-growing social media is becoming increasingly challenging, especially as fine-grained new concepts with defining powers are emerging on social media everyday.
For example, while people are familiar with notions like ``best cars to buy under $50$k'', nowadays people search for ``best electric cars under $50$k''.
Techniques have been developed to perform query and short text conceptualization \cite{kim2013context,wang2015query,song2015open} with topic models and probabilistic graphical models.  However, these methods aim to tag short text or queries which usually consist of several words, and the focus is on word \textit{disambiguation} in the limited context. For example, the query ``most dangerous python in the world'' implies the concept of a \textit{snake} instead of a \textit{programming language}. ``Apple and iPad'' implies a \textit{firm} instead of \textit{fruit}.
Yet, what these techniques cannot solve is to find the defining concepts conforming to human intuition that just characterize the main idea behind the sentence. For example, ``poisonous snakes'' should be a more proper concept instead of just ``snakes'' in the above example.

Furthermore, social media content allows user-created content and usually implies more interesting information than what is available in Wikipedia or conventional knowledge bases, e.g., Probase \cite{wu2012probase}, DBPedia \cite{auer2007dbpedia}, CNDBPedia \cite{xu2017cn}, YAGO \cite{suchanek2007yago}.
For example, a post about ``Mr. Bean has a PhD degree'' naturally reminds a human of \textit{highly educated comedians} or \textit{high IQ actors}, whereas a current knowledge base may only label it with ``British actors'' or ``comedians'', which are correct but not matching the interests of social media viewers. Neither can they help improve recommendation of similar items in mobile apps.
The issue is that the ``British'' and ``Comedian'' parts of Mr. Bean
 are not what the sentence is trying to emphasize in the social media post.
Unfortunately, terms like ``highly educated actors'', although interesting to social media, cannot be found in most existing web-scale knowledge bases.

The Mr. Bean example points to another major issue of most existing knowledge representation methods. Although trying to find relevant concept tags, they do not attempt to understand the meaning of the natural language sentence and make sure the concept to be tagged conforms to the idea of the sentence. 
Taking the case in Figure \ref{case} as an example, the text is talking about a recipe for cooking lotus root. The dish is delicious and can help you ``get rid of hot-and-dry feeling in winter'', which means it is good for health.
However, a common pitfall is to label the content with ``Delicious recipes'', only because it contains words like ``crispy'' and ``fragrant'' that are directly related to ``Delicious recipes'', without respecting the meaning of this specific sentence. In fact, a correct tag to the document should be  ``Healthy recipes'', since the text emphasizes that the dish is healthy.

In this paper, we revisit text conceptualization beyond simple fact-based tagging, disambiguation and topic modeling, by focusing on finding the most accurate concept to characterize the main idea of a text snippet in open-domain social media. Such concept terms must be interesting to online viewers and are thus potentially highly diverse and involving social media slangs. 
We ask the question---can machines comprehend the logical idea behind a piece of social media text and tag it with a \textit{concept} that has just the right defining power and conforms to human intuition? 
According to Kant, ``in order to make our mental images into concepts, one must thus be able to compare, reflect, and abstract.'' (from Logic \cite{LogicKant})
In other words, although it is easy to tag a sentence with relevant phrases, e.g., fruit, animal, trees, yet it is challenging to find the interesting concept at the right granularity that differentiates the sentence from other text mentioning the same entities.

Toward solving this challenge, in this paper, we introduce \textbf{TAG}, a new dataset consisting of $10,000$ concept-sentence pairs obtained from Chinese social media content. Each pair has a binary label indicating whether a (fine-grained) concept term can characterize the logical idea that a sentence wants to convey. The ontology that supplies these concept terms in our dataset is a concept graph, containing social media concepts such as ``best mobile games for girls'', ``top 10 kungfu stars'', created from vast search queries and search logs in QQ Browser, a social media app that serves more than 200 million active users, providing functions such as search, news feeds, etc.
The concept graph organizes a large number of candidate concepts as well as entities in a big graph, where a higher-level concept is termed \textit{superordinate} and a lower-level concept termed \textit{subordinate}. In other words, different concepts and entities are linked by the \textit{isA} relationship in the concept graph.
For example, the concept ``Marvel Superheroes'' or ``The Avengers'' is instantiated by the entities like ``Captain America'', ``Iron Man'', and ``Black Widow''. Furthermore, ``The Avengers'' is a \textit{subordinate} of ``Marvel Superheroes''. 

We evaluate a wide range of latest knowledge representation and neural text matching models on TAG, mainly including sequence-to-sequence text matching methods, e.g., ARC-I, ARC-II \cite{hu2014convolutional}, BiMPM \cite{wang2017bilateral}, MV-LSTM \cite{wan2016deep}, MatchPyramid \cite{pang2016text}, CONV-KNRM \cite{dai2018convolutional}, pre-trained language models, e.g., BERT \cite{devlin2018bert}, and graph embedding methods based on graph neural networks (GNNs) \cite{kipf2016semi}.
However, these state-of-the-art neural network models do not perform particularly well on this social media tagging task, due to their inability to either incorporate contextual information in the concept graph or accurately understand the fine logic conveyed by the sentence. 

Motivated by the challenge of social media tagging in TAG, we further propose a novel \textbf{Graph-Graph matching} framework which converts a pair of concept and sentence to be matched into an aggregated \textit{heterogeneous graph}, by converting the natural language sentence of interest into a graph via syntactical dependency parsing, while incorporating the surrounding connected terms via the concept graph. This allows us to reason over the constructed heterogeneous graph via Relational Graph Convolutional Network \cite{schlichtkrull2018modeling} and acquire a deeper understanding of the logic interactions between terms in the sentence and their level of matching to the candidate concept. 

Furthermore, the proposed Graph-Graph matching method can also tag a new concept which has never appeared in the dataset to a sentence, enabling the \textit{zero-shot learning} ability. This is achieved by inferring the semantic meaning of a new concept based on its subordinate/superordinate relationships to existing concepts in the concept graph. 

Extensive experiments on TAG have demonstrated the superior ability of the proposed model on the social media tagging task over a wide range of state-of-the-art neural matching and language model baselines in terms of accuracy, efficiency, and transferability to new concepts.
To facilitate future research about confining and accurate conceptualization of open-domain social media content, we open source both the TAG dataset and the developed models at \href{https://github.com/XpastaX/TAG}{\ACMDarkBlue{https://github.com/XpastaX/TAG}}.
Our dataset is being constantly updated as new concepts and relationships are being discovered from online social media.

%% file: section/related.tex
\section{Related Work}
\label{sec:related}


\textbf{Text tagging}. Text tagging aims to assign labels to text objects, including documents, sentences and phrases, to represent their main topics. 
Existing studies mainly focus on document tagging \cite{chen2017doctag2vec,liu2019user}, 
where a variety of approaches solve it as a content-based tag recommendation problem for information retrieval \cite{chirita2007p,song2011automatic,rendle2009learning}. 
However, \citet{chen2017doctag2vec} uses predefined tags thus fails to consider the growing tag number in the real-world social media. \citet{liu2019user} focuses on relation mining but fails to utilize the contextual information of its concept tags. Others \cite{chirita2007p,song2011automatic,rendle2009learning} generate tags for a document by combining entities/keywords which can not represent the main idea of the document and lack a higher-level understanding of the text. 


From the machine learning perspective, document tagging can be considered as a multi-label learning (MLL) problem, for which a lot of algorithms have been developed, including compressed-sensing based approach \cite{hsu2009multi}, ML-CSSP \cite{bi2013efficient}, LEML \cite{yu2014large}, FastXML \cite{prabhu2014fastxml}, embedding-based DocTag2Vec \cite{chen2017doctag2vec}, etc.
\citet{liu2019user,liu2020giant} propose a probabilistic inference-based approach and a matching-based approach to tag correlated concepts to a document through the key instances and their contextual words in the document.
There are also studies about concept tagging for sentences, mainly through the perspective of a structured learning problem \cite{gobbi2018concept}.
Algorithms have been developed from both generative and discriminative perspectives, such as LSTM-CRF \cite{huang2015bidirectional,chalapathy2016bidirectional} and Seq2Seq models \cite{sutskever2014sequence}.
However, none of them takes the links between concepts into account, while TAG is the first dataset for tagging that considers them.
    
Lastly, text conceptualization \cite{song2011short,kim2013context,wang2015query,song2015open} aims to link a word or a short phrase to a set of concepts in an ontology or knowledge base, such as Probase \cite{wu2012probase}. 
However, the main tasks of those works focus on the disambiguation of words in the text. In contrast, the concept tagging task in TAG is different in that it focuses on understanding the main idea and semantic meaning of a sentence, when judging whether a concept phrase can be associated with the sentence. The solutions of the above works are mostly based on probabilistic inference and entity-based linking, and the knowledge bases they utilized are mostly constructed from Wikipedia and web pages written in the author perspective which can not represent the interests of general users.
In comparison, our proposed method matches the concept and the sentence by gathering the contextual information of the concepts mined from real-word search logs, which could best preserve the interest and the attention of the users.
 


\textbf{Neural Text matching}. A wide range of neural architectures have been designed for text matching.
Representation-focused models, such as ARC-I \cite{hu2014convolutional} and BiMPM \cite{wang2017bilateral}, transform a pair of text objects into context representation vectors through different encoders and output the matching result based on the context vectors \cite{wan2016deep,liu2018matching,mueller2016siamese,severyn2015learning}.
Interaction-focused models, such as MVLSTM \cite{wan2016deep}, MatchPyramid \cite{pang2016text} and ARC-II \cite{hu2014convolutional}, extract the features of all pair-wise interactions between words in the sentence pair, and aggregate the interaction features through deep neural networks to yield a matching result.
Pre-trained language models such as BERT \cite{devlin2018bert} can also be utilized for text matching through fine-tuning. 
Although neural text matching can characterize the semantic meaning of sentences to a certain degree, these models do not utilize the abundant relational and hierarchical information among concepts.

In contrast, the TAG dataset has called for solutions to a new challenge to the community. It asks the question of how to match a concept-sentence pair by learning from the comparisons between all semantic units in the given sentence and those in the given concept as well as in its subordinate/superordinate concepts. 
Another challenge is how to generalize a trained model to new concepts that have not appeared in training.
There are also works utilizing graph structures \cite{liu2019matching,nikolentzos2017shortest,paul2016efficient} for text matching. However, by evaluating a wide range of techniques including GNN-based methods, we observe that the proposed concept tagging problem cannot readily be solved by combining existing off-the-shelf methods.

\textbf{Concept graph construction}. The concept graph is constructed to empower models with the knowledge about the concepts in the real world. Existing works mainly focus on extracting the relations between entities in a text and concepts \cite{ji2019microsoft,liang2017probase+,wang2016understanding}. Instead, our main task is to conceptualize a document by tagging the sentences in the document with the most semantically relevant concepts, aided by the concept graph directly sampled from Attention Graph \cite{liu2020giant}.

%% file: section/data.tex
\section{The TAG Dataset}
\label{sec:data}

\begin{figure*}[htp]
	\centering
		    \includegraphics[width=0.98\textwidth]{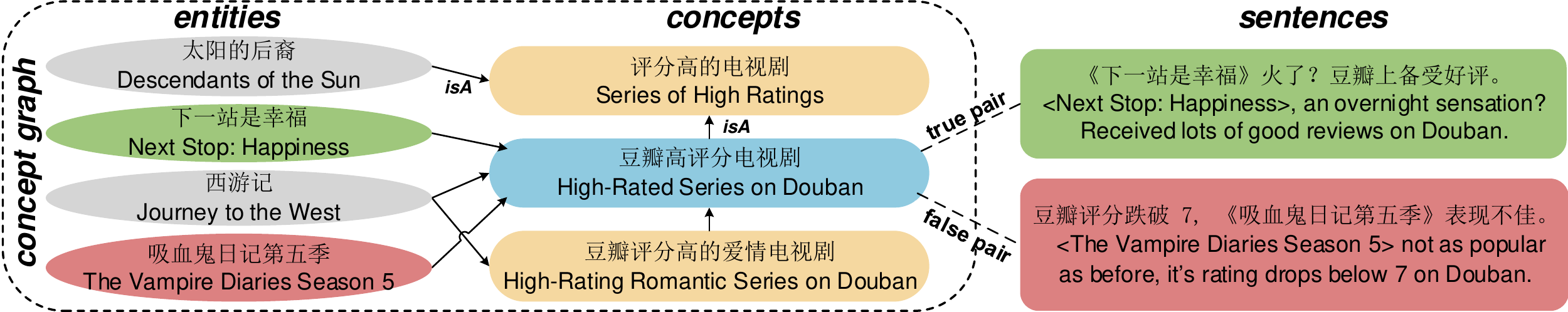}
		\caption{The TAG dataset consists of concept-sentence pairs with binary labels, depending on whether the concept can be concluded from the given sentence or not. The above is a snippet of the concept graph related to a concept ``Highly-Rated Series on Douban" together with two sample pairs. \textit{isA} means "is a subordinate of". }
	\label{data_sample}
	\vspace{-3mm}
\end{figure*}

In this section, we formally introduce the concept tagging problem as a concept-sentence matching problem in the context of a concept graph. We describe the procedure in which the TAG dataset is constructed and analyze its unique characteristics. 

The TAG dataset is constructed from the vast search logs in QQ Browser. It is collected and formed in a concept matching style since the developed methods are cheap and easy to be integrated and transferred to real world applications, compared with other advanced research tasks such as ranking or text generation. For example, tagging a concept to a newly uploaded document to alleviate the cold-start problem in recommendation or improve its summarization.

To emphasize the importance of zero-shot and few-shot learning in real-world scenarios, we gathered and developed solutions on a medium-sized dataset with a large number of concepts. Specifically, the dataset contains two parts: i) 10,000 concept-sentence pairs, each with a binary label indicating whether the concept can characterize the meaning of the sentence and thus be tagged onto the sentence; ii) a concept graph that contains a collection of phrases mined from vast amounts of online social media content, representing \textit{concepts}, and \textit{entities}, as well as the \textit{isA} relationships between them.
Specifically, the concepts are mined from search logs where popular queries satisfying certain rules serve as the concepts. The concept graph used in the TAG dataset is sampled from the \textit{Attention Graph} constructed in GIANT \citet{liu2020giant}. However, the same tagging techniques apply to other ontologies or knowledge bases, such as Probase \cite{wu2012probase}, DBPedia \cite{auer2007dbpedia}, AliCoCo \cite{luo2020alicoco}, etc.

As a user-centered ontology, Attention Graph is constructed from a web-scale corpus and is able to abstract entities and events into the higher-level concepts and topics based on user interests.
Most existing taxonomy or knowledge bases, e.g., Probase, DBPedia, CNDBPedia  \cite{xu2017cn}, YAGO \cite{suchanek2007yago}, utilize Wikipedia as their main source of content, which are not the best at tracking user-centered phrases on the online social media. 
Instead, the Attention Graph is mined from the \textit{click graph}, which is a large bipartite graph formed by search queries and their corresponding clicked documents that can best represent the attention and interests of online users. 
For example, if a query is ``The academic degree of Mr. Bean'', many knowledge bases may recommend related results such as the movies of Mr. Bean. Attention Graph is able to infer the true interest of the user, which is in ``highly educated comedians'' and gives search results about other comedians who have high academic achievements.
More details regarding the construction of this concept graph can be found in \citet{liu2020giant}.
In the concept graph we utilize, each concept is a phrase that represents a collection of entities sharing some common properties.
The major relational connections in the concept graph are the \textit{isA} (subordinate) relationships between the terms.

\subsection{Task Description}
The concept tagging task considered here is a challenging new problem that rarely tackled in prior research.
First, in TAG, each concept only appears in a few training sample pairs, leading to a few-shot learning problem, as opposed to traditional text classification where each class has many training samples. Second, each concept is usually composed of a few words and thus contains limited semantic information. 
Third, in real-world social media, new concepts constantly emerge on a daily basis, requiring the tagging model to generalize and transfer to new concepts that have not appeared in the training set. 

We argue that a critical component to support the few-shot learning and transferability of a tagging model is to utilize the relational information in the concept graph, i.e., the comparisons among concepts and links between concepts and entities.
To the best of our knowledge, TAG is the first dataset specifically designed for online social media content tagging based on graphical knowledge representations. 
Existing datasets for document tagging, e.g., the UCCM dataset \cite{liu2019user}, do not have a concept graph component which lays the relational context around concepts and entities.

\subsection{Dataset Construction}

Table~\ref{tab_stat} presents the statistics of the TAG dataset. TAG contains 10,000 data examples with 230,504 relations and 5,438 concepts.
Such a large amount of tags poses a big challenge to traditional labeling and text matching models. Figure~\ref{data_sample} illustrates two concept-sentence pairs, a positive pair (matched) and a negative pair (unmatched). Two sentences on the right both contain some entities (titles of TV series) that belong to the concept ``Highly-Rated Series on Douban'' in the concept graph. However, the second sentence criticizes the Vampire Diaries and, together with the concept, should be labeled as a negative pair. We describe the detailed procedures of constructing the TAG dataset in the following:

\textbf{Candidate set retrieval.} 
We sampled 5438 concepts and their corresponding contexts only from the ``entertainment'' domain of Attention Graph \cite{liu2020giant} to form our concept graph.
To prepare candidate concept-sentence pairs for human labeling, for each selected concept, we record the top-$3$ clicked documents retrieved by inputting the concept as a search query. For each document, we select the sentences which contain at least one entity that has \textit{isA} relationship with the query concept. 
Note that even if a sentence contains an entity that belongs to a concept, this concept does not necessarily characterize the main idea of the sentence.
Furthermore, many concepts share similar meanings as they are retrieved from similar search logs. For example, ``Highly rated movies'' has the same meaning as ``Movies that receive a high rating''. Besides, many concepts are actually the combinations of other concepts. For example, the concept ``Action movie with a high rating'' is a combination of  ``Action movies'' and ``Highly rated movies''. 
To enhance the quality and cleanliness of the TAG dataset which aims to facilitate the research in concept tagging, we further increase the diversity of the concepts in TAG.
We use a filter to prevent semantically redundant concepts from being tagged to the same sentence. Specifically, for all the concepts paired to a sentence in the dataset, we rank them by the summation of frequencies of words and keep at most 4 concepts with the lowest aggregate word frequencies. This will also remove obvious matches to popular and easy concepts and increase the interestingness of the concepts.

\input{table/datafeatures}

\textbf{Labelling concept-sentence pairs.} 
All of our examples are from the ``entertainment'' domain, which is not deep or technical. Thus the annotation only requires common sense. We employ three annotators to ensure the quality of our dataset.
Given the candidate pairs retrieved based on entities, each concept-sentence pair will be labeled with either 1 or 0 by two annotators (who are hired professional data labeling personnel yet totally ignorant and neutral to the research work being conducted), by solely examining the concept-sentence pair based on their own understanding without external information.
If the results from the two annotators are different, the third annotator will be hired for a final check.

The criterion of the annotation is based on whether the concept can be concluded from the given sentence. In Figure~\ref{data_sample}, the sentence ``<Next Stop: Happiness>, an overnight sensation? Received lots of good reviews on Douban'' forms a positive example with the concept ``High-rated series on Douabn'' since it ``received lots of good reviews on Douban''. In contrast, the sentence ``<The Vampire Diaries Season 5> not as popular as before, its rating drops below 8 on Douban'' is a negative pair with the concept since the series is ``not popular as before'', which contradicts the concept.

\begin{figure*}[htp]
	\centering
		    \includegraphics[width=0.95\textwidth]{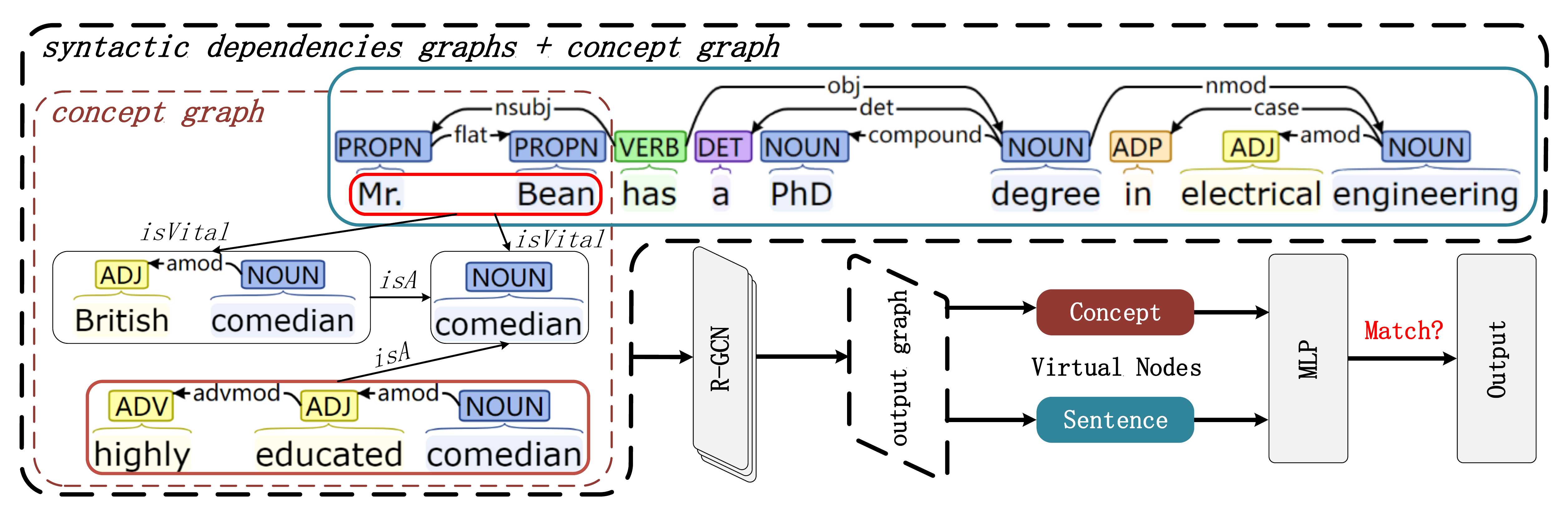}
		\caption{The procedure of predicting an example. Graph circled by the black dash line is the heterogeneous graph constructed for a concept-sentence pair, with the context information from the concept graph.}
	\label{graph_structure}
	\vspace{-3mm}
\end{figure*}
\textbf{Post-processing.} After human labeling, the dataset is actually unbalanced---there are more negative examples than positive ones. We down-sample the negative pairs to ensure that the positive and negative examples are balanced in the dataset. 
Furthermore, to simulate the real word scenario where the number of concepts keeps growing, we split the 10,000 sample pairs in TAG for training and testing under two different strategies: i) random splitting; ii) non-overlapped splitting, where the latter ensures the concepts that have appeared in the training set do not appear in the test set at all. The non-overlap scenario can be used to test the \textit{zero-shot learning} capability of a model, i.e., the ability to generalize to new concepts it has never seen. 


%% file: table/datafeatures.tex
\begin{table}
\caption{The statistics of the TAG dataset. \textit{CG rel.} denotes the number of relations in the concept graph.}
\begin{tabular}{l|c|cc|cc}
\bottomrule[1pt]
            &&&&&\\[-2ex]
             & \multirow{2}{*}{Total} & \multicolumn{2}{c|}{Non-Overlap} & \multicolumn{2}{c}{Random} \\[0.3ex] \cline{3-6} 
             &&&&&\\[-2ex]
             &                        & Train              & Test             & Train            & Test           \\ \hline
             &&&&&\\[-2ex]
Concepts      & 5438                   & 4326               & 1112             & 4660             & 1594           \\[0.3ex]
Sentences     & 8085                   & 6780               & 1900             & 6733             & 1901           \\[0.3ex]
Positive    & 4152                   & 3380               & 772              & 3332             & 820            \\[0.3ex]
Pairs    & 10000                  & 8000               & 2000             & 8000             & 2000           \\[0.3ex]
CG rel. & 230504                 & 188312             & 42192            & 181461           & 49043          \\ \toprule[1pt]
\end{tabular}
\label{tab_stat}
\vspace{-5mm}
\end{table}

%% file: section/model.tex
\section{Models}
\label{sec:model}
In this section, we propose a novel and efficient solution for our concept-sentence tagging problem, which is a Graph-Graph matching model based on the representation of the given concept-sentence pair in a \textit{heterogeneous graph}. 
Before that, we first describe two existing neural text matching approaches that can also be applied to the problem, and show their limitations.

\subsection{Seq-Seq and Graph-Seq Matching} 
There are two existing solutions that are intuitive and applicable to our tagging task.
The first one is to build a sequence to sequence (Seq-Seq) matching model to leverage the semantic information contained in the concept-sentence pair.  
Specifically, we fine-tune the BERT \cite{devlin2018bert} model to test sequence matching without any context information for the concept. We encode both the given concept and sentence by BERT, then either use the \text{[CLS]} tags or the embeddings of the sequences (BERT outputs one embedding for each Chinese character of the input text, and we encode those outputs by a Bi-LSTM layer) to perform binary classification. Denote the output embeddings of the sentence and the concept as $V_S$ and $V_T$, respectively, we pass $[|V_S - V_T|, V_S \circ V_T]$ into a Multi-layer perceptron (MLP) classifier to make the prediction.

Another natural solution to the proposed task is to build a Graph-Sequence (Graph-Seq) matching model to encode the contextual information of the given concept in the concept graph using a GNN. Specifically, on the graph side, the local context related to the concept is mapped into a directed graph with each node of the entities and the concepts represented by its word vectors (or the mean of the word vectors for the concepts formed by multiple words), and fed into a Relational Graph Convolutional Network (R-GCN) \cite{schlichtkrull2018modeling} to aggregate the contextual information in the concept graph. On the sequence side, we calculate the mean of the word vectors in the sentence as its embedding. Then, with the same operation performed in the Seq-Seq matching model, the node embedding of the concept is combined with the embedding of the sentence, and sent to an MLP classifier for prediction. 
We will introduce the details of the R-GCN model later in this section.

Both of the above two methods have some limitations. The Seq-Seq matching model fails to leverage the rich contextual information provided by the concept graph, while the Graph-Seq matching model lacks sufficient semantic support. Although the contextual information of the concept is utilized by the Graph-Seq model, the potential connection between the concept and the sentence is not explored. In other words, how the concept could relate to the sentence is not considered in the model.


\subsection{Graph-Graph Matching} 


To incorporate the context information in the concept graph and model the 
interactions within a concept-sentence pair, we propose to form a \textit{heterogeneous graph} representation of the given concept-sentence pair and combine it with R-GCN to estimate whether the concept is coherent with the main content of the sentence.

\begin{figure}[tp]
	\centering
		    \includegraphics[width=0.9\columnwidth]{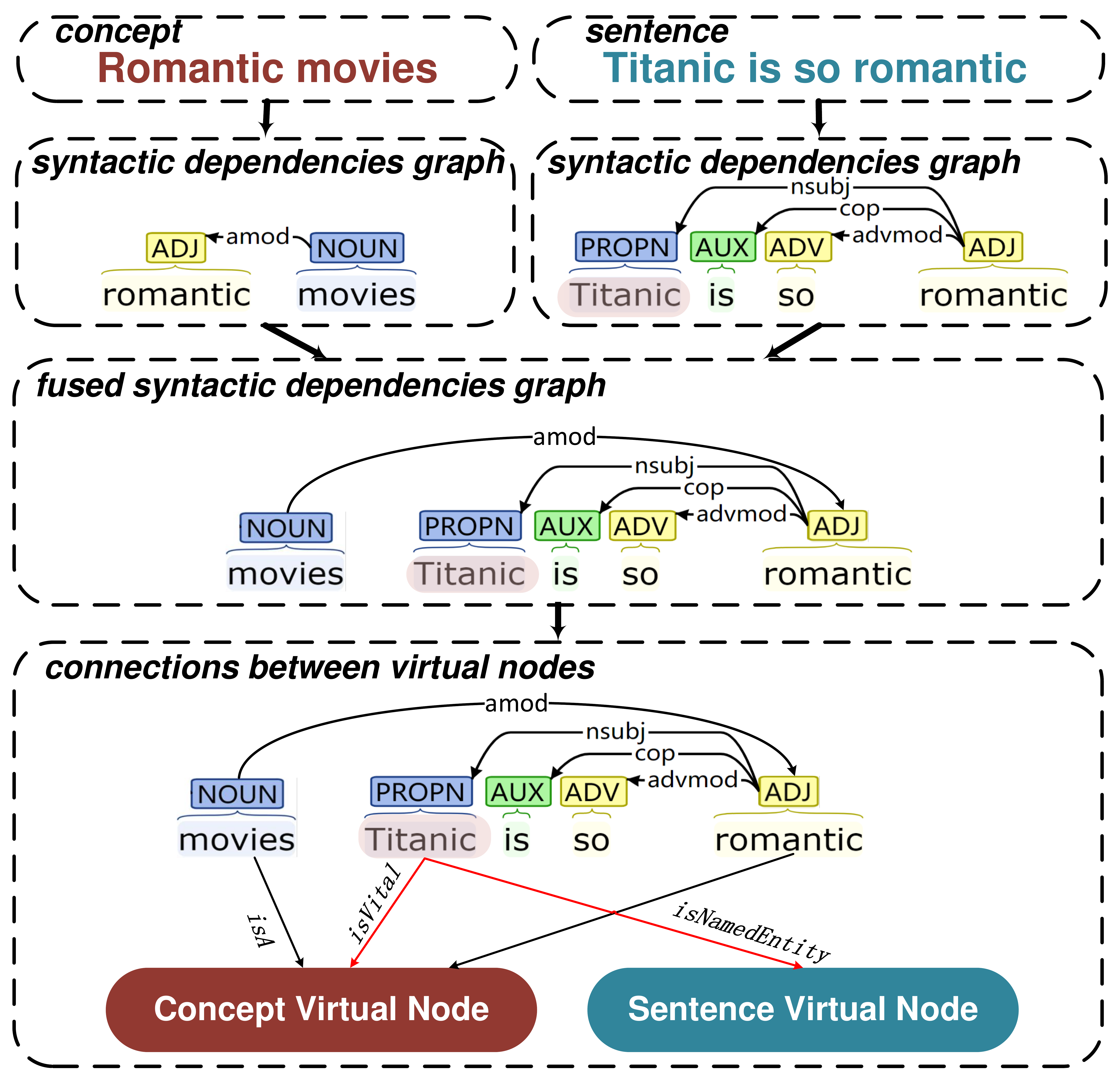}
		\caption{The process of connecting the concept and the sentence. ``Titanic'' is the entity shared between the concept graph and the sentence.}
	\label{denpendency_graph}
	\vspace{-5mm}
\end{figure}

\textbf{Heterogeneous graph construction.} 
Figure \ref{graph_structure} presents the heterogeneous graph structure of an example pair.
For each concept-sentence pair, denote the sentence by $S$ and the concepts in context by $C=\{C_T, C_C\}$, where $C_T$ is the concept paired with the sentence (circled in red) and $C_C$ (circled in blue) represents other neighboring concepts in the concept graph. We further denote the set of entities associated with $C$ as $E_C$ and set of named entities in the sentence $S$ as $E_S$. Next, we introduce our procedure for constructing the heterogeneous graph based on Figure \ref{denpendency_graph}.

For each example, the first step is to obtain all syntactic dependencies in $S$ and $C$. 
As presented in Figure \ref{graph_structure}, the sentence and the concepts will be segmented and fed into the Stanford CoreNLP model \cite{manning2014stanford,qi2020stanza} to generate a set of syntactic graphs in Figure \ref{denpendency_graph}. 
Each word is a node in these dependencies graphs either for the sentence or the concepts.
To simplify the graphs, we fuse those syntactic graphs together by merging all nodes of the same word, as presented in Figure~\ref{denpendency_graph}.
The fused heterogeneous graph $G_{syn}$ then stores all semantic information. 

Second, we create \textit{virtual nodes} for each element in $S\cup C$ in the graph and establish connections between those virtual nodes and $G_{syn}$. 
Specifically, we link the concept nodes and all of its words together with \textit{isA} relation, then we link the sentence node and all of its named entities together with \textit{isNamedEntity} relation. 
With such connections, the virtual nodes perform like hubs, which aggregate all semantic information among $G_{syn}$.

Finally, to utilize the concept graph, for each entity in $E_C\cap E_S$ and each concept in $C$, we map their context relations into the heterogeneous graph (red dash circle in Figure~\ref{graph_structure}). Moreover, to emphasize the entities shared by $S$ and $C$, we assign the links between such entities and $C$ with a special relation named \textit{isVital}. Since the concept-sentence pair is retrieved by those shared entities, such links are expected to help distinguish a successful retrieval.

There are in total 36 types of dependencies and 3 types of manually introduced relations in the heterogeneous graph.
Note that the relations in the graph are all directed. Thus, some nodes may not have outward links to pass their messages to others. To fix this, for each relation in the heterogeneous graph, we introduce an opposite relation to ensure information transmission between the nodes, which doubles the number of relations to 78. This fix is also taken in the Graph-Seq matching model.

\textbf{Relevance estimation.} As shown in Figure~\ref{graph_structure}, we apply R-GCN on the heterogeneous graph of the concept-sentence pair to encode the virtual nodes of $S$ as $V_S$ and $C_T$ as $V_T$. Then we pass $[|V_S - V_T|, V_S \circ V_T]$ into an MLP classifier to make a binary decision on the concept-sentence pair. 
Note that the concept and the sentence are embedded together, which means the same concepts will have different embedding if paired with a different sentence. The insight is that the concept and the sentence will exchange their information to output the embedding specialized for the pair, which could lead to a better prediction.

The embedding of each node in the heterogeneous graph is formed by the following features: the word vector, part-of-speech (POS) tag, named entity recognition (NER) tag, and the source of the node. Each part will be tagged with ``None'' if not applicable. Specifically, we employ the word vectors from \cite{song-etal-2018-directional}, which contains a large number of domain-specific words, entities, slangs, and fresh words that are of high frequencies in our TAG dataset. For each virtual node in the heterogeneous graph, we use the mean of its word vectors as the embedding. 
For the source tag, we use ``SENTENCE'', ``CONCEPT'' and ``BOTH'' to represent where the node comes from.

\textbf{Relational GCN.} We now briefly describe the  R-GCN used in Graph-Graph and Graph-Seq matching models. Denote a directed graph as $G = (V, E, R)$ with labeled edges $e_{vw} = (v, r, w) \in E$, where $r \in R$ is a type of relation and $v, w \in V$ are two nodes connected by $r$.
In the message-passing framework \cite{gilmer2017neural} of graph neural networks, the hidden states $h^l_v$ of each node $v \in G$ at layer $l$ are updated based on messages $m^{l+1}_v$ according to:
\begin{align}
m^{l+1}_{v} &= \sum_{w\in N(v)} M_l(h^l_v, h^l_w, e_{vw}), \\
h^{l+1}_v &= U_l(h^l_v, m^{l+1}_v),
\end{align}
where $N(v)$ are the neighbors of $v$ in $G$, $M_l$ and $U_l$ are the message function and vertex update function at layer $l$, respectively.

The message passing function of Relational Graph Convolutional Networks is defined as:
\begin{align*}
h_v^{l+1} = \sigma\Bigg(\sum_{r \in R} \sum_{w\in N^r(v)} \frac{1}{c_{vw}} W^l_r h^l_w + W^l_0 h^l_v \Bigg),
\end{align*}
where $N^r(v)$ is the set of neighbors under relation $r\in R$. $\sigma(\cdot)$ is an element-wise activation function such as ReLU$(\cdot) = \text{max}(0, \cdot)$. $W^l_r$ and $W^l_0$ are learned weight matrices. $c_{vw}$ is a problem-specific normalization constant that can be learned or pre-defined (e.g., $c_{vw} = |N^r(v)|$). 
We can see that R-GCN learns relation-specific transformation matrices to take the type and direction of each edge into account.

Moreover, R-GCN exploits basis or block-diagonal decomposition to regularize the number of weights.
For basis decomposition, each weight matrix $W^l_r \in \mathbb{R}^{d^{l+1}\times d^l}$ is decomposed as
\begin{align}
W^l_r = \sum_{b=1}^{B} a^l_{rb} V^l_b,
\end{align}
where $V^l_b \in \mathbb{R}^{d^{l+1}\times d^l}$ are base weight matrices.
In this way, only the coefficients $a^l_{rb}$ depend on $r$. For block-diagonal decomposition, $W^l_r$ is defined through the direct sum over a set of low-dimensional matrices:
\begin{align}
W^l_r = \mathop{\bigoplus}_{b=1}^{B}Q^l_{br},
\end{align}
where $W^l_r = \text{diag}(Q^l_{1r}, Q^l_{2r}, \cdots, Q^l_{br})$ is a block-diagonal matrix with $Q^l_{br} \in \mathbb{R}^{(d^{l+1}/B) \times (d^{l}/B)}$.
The basis function decomposition introduces weight sharing between different relation types, while the block decomposition applies sparsity constraint on the weight matrices.

%% file: section/simu.tex
\section{Experiments}
\label{sec:exp}
\input{table/result}
\input{table/case}
In this section, we evaluate the proposed approach for concept tagging on TAG with a number of state-of-the-art neural text matching baselines.

\textbf{Baselines and Metrics.} We compare the proposed method with 6 existing baseline methods, including ARC-I, ARC-II \cite{hu2014convolutional}, BiMPM \cite{wang2017bilateral}, MV-LSTM \cite{wan2016deep}, MatchPyramid \cite{pang2016text}, CONV-KNRM \cite{dai2018convolutional}, and 3 designed baseline methods: Graph-Seq matching, BERT-Siamese and BERT-CLS. The 6 existing baselines and the BERT model all perform semantic matching of a given concept-sentence pair, while the Graph-Graph matching and Graph-Seq matching models also leverage the contextual information in the concept graph. We use the Chinese word vector proposed in \cite{song-etal-2018-directional} as the common input to all the models except BERT.

The performance of each model is evaluated on the two versions of TAG, including random and non-overlapped versions, as shown in Table~\ref{tab_stat}. We utilize 1,000 sample split from the training set for validation. Additionally, for the \textit{non-overlapped} training set, we will ensure the concepts in the validation set are different from those remains in the training set as well. The evaluation metrics are the F1 score and the accuracy of the prediction.

\textbf{Implementation details.} The 6 existing baselines are implemented by MatchZoo \cite{Guo:2019:MLP:3331184.3331403}. The BERT is from Transformers \cite{Wolf2019HuggingFacesTS}. And our graph models are developed with Pytorch Geometric \cite{Fey/Lenssen/2019}. 
We have manually tuned the hyper-parameters of all the models.
For the \textit{Graph-Graph matching} model, 
we set the number of R-GCN layers to 3 and the number of bases to 14. The initial learning rate is $1e^{-4}$ and with a cosine decay scheduler. The model is trained for 20 epochs with 10\% steps for warm up.
For the \textit{Graph-Seq matching} model, We use a 3-layer and 2-base R-GCN on the graph side, since we only have \textit{isA} and its opposite relation. 
The learning rate and scheduler are the same as those in Graph-Graph matching. The number of epochs is 50.
For \textit{BERT}, we fine-tune the model with a batch size of 8 and a learning rate of $1e^{-6}$ for 30 epochs with 10\% steps to warm up.
For each model, the epoch with the best validation F1 score is used for final evaluation.

\subsection{Results and Analysis}

Table~\ref{tab_result} summarizes the performance comparison of all the methods on the two versions of TAG. Each score in the table is an average of 5 runs. 
We can observe that the proposed Graph-Graph matching method significantly outperforms all other baselines.
Comparing the Graph-Seq matching model with the 6 existing baselines (i.e., from ARC-I to CONV-KNRM), the Graph-Seq matching model outperforms all of them. That confirms our hypothesis that the relational information in concept hierarchy offers additional help to the prediction. 
Furthermore, the proposed Graph-Graph matching model significantly outperforms the Graph-Seq matching model, since the syntactic graph provides richer semantic information than encoding a concept/sentence by taking the mean of word vectors.

BERT outperforms all other baseline methods because it has learned lots of prior knowledge through pre-training. Our proposed model outperforms the BERT methods with the 40 times smaller model size: both of the BERT implementations has more than 110 million parameters while our proposed method contains less than 2.66 million parameters. This is because the prior knowledge in BERT is implicit and might be inaccurate. The Chinese BERT model is trained based on Chinese characters, while each word is usually formed by at least two characters. 
Such a training method will lead to misunderstanding when fresh words or new slangs show up in the sentence. 
Also, the training corpus of Chinese BERT is mainly collected from official sites (e.g. Wikipedia) which are usually written in an objective aspect, while in real case application, most online texts are written in user-aspect containing complicated sentiments and willing.
Different from BERT, with proper word segmentation, the proposed method utilizes the word vectors trained with a large amount of user-aspect corpus that contains lots of slang and self-made words. Therefore, it can better understand the text.
Furthermore, BERT does not utilize the context information in the concept graph either.
Compared with BERT, the syntactic graph provides great advantages to semantic aggregation. With the additional context contained in the concept graph, the model can characterize the semantics of the new concepts more accurately, leading to better performance. 

Our model also gives better results on Non-overlapped data because each new concept will be linked with the sentence through the shared entities retrieved from the concept graph, e.g., ``Titanic'' in Figure~\ref{denpendency_graph}.
The contextual information in the concept graph can further help in understanding the new concept. By leveraging both the advantages, our Graph-Graph matching model enriches the information contained in the output embedding, which helps its MLP classifier performs better prediction on zero-shot learning.

\subsection{Ablation Study}

Furthermore, to measure the impacts of the syntactic dependencies graph (SDG) and the concept graph (CG) to our proposed model, we perform ablation studies on our proposed method by removing the SDG and the CG, respectively. As shown in Table~\ref{tab_result},
\textit{``G-G\ w/o\ Syn''} represents the model without syntactic dependencies graph, while \textit{``G-G\ w/o\ CG''} represents the model without the concept graph.
Both variations yield lower performance compared to the original one. 
Our proposed method of extracting information from SDG reaches a higher performance than BERT. The \textit{``G-G\ w/o\ Syn''} model can also reach a comparable performance with a 100 times smaller model size than BERT methods.

Compare the impacts of CG and SDG on the proposed model, SDG has a greater overall improvement on the performance of the model. Together with the word vectors, our proposed method can extract rich semantic information from the syntactic dependencies. The performance of \textit{``G-G\ w/o \ CG''} model decreases less than \textit{``G-G\ w/o \ Syn''}, this is because the SDG is the backbone of extracting semantic information, which is most important for understanding a sentence. Instead, CG mainly contributes to providing contextual information to enhance the understanding of a sentence, especially for zero-shot learning.
By putting CG back to the \textit{``G-G\ w/o \ CG''} model, the F1 score increases 3.04\% for the non-overlapped dataset while increases 1.66\% for the random dataset. This fact further shows the importance of CG in TAG: the contextual information it provides can help the model better understand the new concept, which leads to higher performance on the zero-shot tagging task.

From the results above, we can conclude that both of the contextual information in CG and the syntactic dependencies contribute to improving the semantic understanding. The contextual information in CG of TAG shows excellent potential on the zero-shot task. Our proposed method well unitizes both of the advantages to achieve superior performance.

\subsection{Error Analysis}

To analyze the weaknesses of both the proposed model and the BERT model, we further gathered all the incorrect predictions of the non-overlapped dataset. 
As shown in Table~\ref{tab_case}, we can generally categorize the reason for the errors into three main types: lack of reasoning ability, failed Named Entity Recognition, and complex context.
Our model is less prone to the first type of errors in general, and the last two types are only common in the BERT model.

\textbf{Reasoning ability.} 
Words may have implicit information that is hard to extract. In the first example in Table~\ref{tab_case}, the concept is about the \textbf{movie} of ``Resident Evil 7'', while the sentence is talking about the \textbf{game} ``Resident Evil 7''.
The annotators label the example as ``False'', because ``speed run'' means clearing a game as fast as possible and ``Zoe ending'' suggests multiple endings exist, while the movies usually tell stories with single-endings.
However, the models may lack the related knowledge to perform such reasoning and make the correct prediction.
As shown in this example, BERT incorrectly predicts the label as ``True'' since it failed to distinguish ``Resident Evil 7'' in the sentence from ``Movie Resident Evil 7'', while our proposed method successfully recognizes ``Resident Evil 7'' in the sentence as a video game and predict the example as ``False''.
For all 14 concept-sentence pairs with the concept ``Movie Resident Evil 7'' in the test set, BERT made 13 false predictions while our model made only 4 false predictions. 
However, by comparing the difference between the entities in the concept graph and the sentence, e.g. the entity ``Zoe'' is related to the game ``Resident Evil 7'' while not exist in the movie, we may further improve the performance of our proposed model.

\textbf{Named Entity Recognition.} Limited by the nature of the Chinese language, BERT will encode a Chinese sentence in characters. This could cause problems when it comes to embedding an entity, which usually is a combination of multiple characters. Take the second sentence in Table~\ref{tab_case} as an example, Gary Copper is the name of a person, which in Chinese is ``\begin{CJK}{UTF8}{gbsn}加里·库柏\end{CJK}'', formed by 4 unrelated Chinese characters since it is a transliteration of a name.
BERT is not able to distinguish such entity and the irrelevant characters make it more difficult to understand the sentence. It then outputs an inaccurate embedding of the sentence, which leads to a false prediction. 
In contrast, our proposed model utilizes the NER technique to distinguish those entities, extracts the meaning of them from word vectors, and output the correct embedding. 

\textbf{Complex context.} Long sentences may have complex and rich information which could be hard for the models to understand. As an example, in the third sentence of Table~\ref{tab_case}, the context information includes: ``a reader called the hotline of the newspaper'', ``the reader has been following the TV series Behind'', and ``the plot about the hospital in Behind is true''. Based on these information, we can draw the key ``Behind is a TV series about hospitals'' and match the sentence with the concept ``TV series about hospitals''.
However, when encoding the sentence, BERT will not focus on ``Behind is a TV series about hospitals''. The key information is then ``diluted'' in the output embedding, which results in the mismatch between the concept and the sentence. 
In contrast, our proposed method successfully established the connections between the concept and the sentence as described in Section~\ref{sec:model} and integrated the connections in encoding.
Specifically, in this example, the concept and the sentence are connected by the words ``TV series'', ``Behind'' and ``hospital'', through which the information will be passed to encode the concept and the sentence. 
Such mechanism emphasizes the significance of the concept during the encoding process, and results in a better prediction in complex context situations.

%% file: table/result.tex
\begin{table}
\caption{Experimental results.}
\begin{tabular}{l|cc|cc}
\bottomrule[1pt]
\multicolumn{1}{c|}{\multirow{2}{*}{}} & \multicolumn{2}{c|}{} & \multicolumn{2}{c}{} \\[-2ex]
\multicolumn{1}{c|}{\multirow{2}{*}{Model}} & \multicolumn{2}{c|}{Overlapped} & \multicolumn{2}{c}{Non-overlapped} \\
\multicolumn{1}{c|}{}                       & F1                 & Acc                & F1                  & Acc                  \\ \hline
&&&&\\[-2ex]
ARC-I                &64.97      &71.48      &56.51      &65.99    \\[0.1ex]
ARC-II               &66.39      &70.50      &56.11      &64.59    \\[0.1ex]
BiMPM               &62.50      &69.66      &58.46      &66.62    \\[0.1ex]
MV-LSTM              &66.60      &71.10      &59.61      &64.19    \\[0.1ex]
MatchPyramid        &63.70      &71.32      &56.87      &67.06    \\[0.1ex]
CONV-KNRM           &66.93      &73.31      &60.59      &64.93    \\[0.1ex] \hline
&&&&\\[-2ex]
G-S                 &68.26      &73.84      &61.89      &67.58    \\[0.1ex]
BERT-Siamese        &69.91      &74.64      &65.15      &70.43    \\[0.1ex]
BERT-CLS            &70.14      &74.49      &65.21      &70.32    \\[0.1ex] \hline
&&&&\\[-2ex]
G-G w/o Syn       &69.14      &74.61      &63.98      &69.72    \\[0.1ex] 
G-G w/o CG        &74.38      &79.06      &66.73      &72.98    \\[0.1ex] 

G-G                 &\textbf{76.04}      &\textbf{80.26}      &\textbf{69.77}      &\textbf{75.72}                \\[0.1ex] \toprule[1pt]
\end{tabular}

\label{tab_result}
\vspace{-5mm}
\end{table}

%% file: table/case.tex
\begin{table*}
\caption{Typical types of errors in concept matching for social media content.}
\begin{tabular}{l|ccl}
\bottomrule[1pt]
&&&\\[-2ex]
\multicolumn{1}{c|}{Error Types}           &Label       & Concept  & \multicolumn{1}{c}{Sentence}    \\[0.2ex]\hline
&&&\\[-2ex]

Reasoning Ability       &False 
& Movie Resident Evil 7   
&``Resident Evil 7'', 45min speed run video of Zoe ending.                  \\[0.1ex]\hline
&&&\\[-2ex]
\multirow{2}{*}{Named Entity Recognition}         &\multirow{2}{*}{True} 
&\multirow{2}{*}{Previous Oscars winners}  
&On the 14th Oscars in 1942, Gary Cooper was nominated for the\\ 
&&&second time and finally won the award.\\[0.1ex]\hline
&&&\\[-2ex]
\multirow{3}{*}{Complex Context}         &\multirow{3}{*}{True} 
& \multirow{3}{*}{TV series about hospitals}   
&Yesterday, a reader who has been following the TV series ``Behind'' \\
&&&called the hotline of our newspaper and said: The current hospital \\
&&&is indeed like that, and the plot is very real.
\\[0.1ex]\toprule[1pt]

\end{tabular}
\label{tab_case}
\vspace{-3mm}
\end{table*}

%% file: section/conclude.tex
\section{Conclusion}
\label{sec:conclude}

In this paper, we present TAG, a high quality labelled dataset of 10,000 concept-sentence pairs to facilitate research on fine-grained concept tagging for social media content understanding in the presence of a concept graph. 
By evaluating a wide range of state-of-the-art neural text matching methods, we observe that the proposed concept tagging problem becomes a challenging task, as existing sequence-to-sequence or graph embedding methods either fail to leverage the context information in the concept graph or cannot accurately grasp the logic meaning conveyed in the natural language sentence. 
The need to generalize and transfer models to new concepts further increases the challenge.
We further design a novel framework to convert text to a graph via dependency parsing such that Graph-Graph matching can be conducted between the sentence and the concept in the context of the concept graph. 
Extensive experiments have demonstrated the superior ability of the proposed method to utilize contextual information in the concept graph, to reason about the logical match between a concept and a sentence, and to generalize and transfer to new concepts.  